\documentclass[letterpaper]{article}
\usepackage{aaai}
\usepackage{times}
\usepackage{helvet}
\usepackage{courier}
\usepackage{graphicx}
\usepackage{multirow}
\usepackage{booktabs}
\usepackage{amsmath}
\usepackage{natbib}
\usepackage{kotex}

\begin{document}
%
\title{External Knowledge Selection with Weighted Negative Sampling in Knowledge-grounded Task-oriented Dialogue Systems}
\author{{Janghoon Han, Joongbo Shin, Hosung Song, Hyunjik Jo, Gyeonghun Kim, Yireun Kim, Stanley Jungkyu Choi}\\
LG AI Research\\
Seoul, Korea\\
\{janghoon.han, jb.shin, hosung.song, hyunjik.jo, ghkayne.kim, yireun.kim, stanleyjk.choi\}@lgresearch.ai\\
}
\nocopyright
\maketitle
\begin{abstract}
\begin{quote}
Constructing a robust dialogue system on spoken conversations bring more challenge than written conversation. In this respect, \texttt{DSTC10-Track2-Task2} is proposed, which aims to build a task-oriented dialogue (TOD) system incorporating unstructured external knowledge on a spoken conversation, extending \texttt{DSTC9-Track1}. 
This paper introduces our system containing four advanced methods: data construction, weighted negative sampling, post-training, and style transfer.
We first automatically construct a large training data because \texttt{DSTC10-Track2} does not release the official training set.
For the knowledge selection task, we propose weighted negative sampling to train the model more fine-grained manner. We also employ post-training and style transfer for the response generation task to generate an appropriate response with a similar style to the target response. In the experiment, we investigate the effect of weighted negative sampling, post-training, and style transfer. Our model ranked 7 out of 16 teams in the objective evaluation and 6 in human evaluation.
\footnote{https://github.com/hanjanghoon/Weighted\_NS}

\end{quote}
\end{abstract}

\section{Introduction}
Task-oriented dialogue (TOD) systems, which aim to assist users with specific tasks through conversation, have received much attention in research and industry due to their applicability in various services such as personal assistants and customer chatbots \citep{chen2017survey}. In general, TOD systems are able to respond to the user based on a given DB or API. In reality, however, a user often requests detailed information that exceeds the DB or API coverage, such as whether a companion with a pet is allowed in a restaurant may not be included in the DB or API.

\begin{figure}[t]
\centering
\includegraphics[width=3.0in]{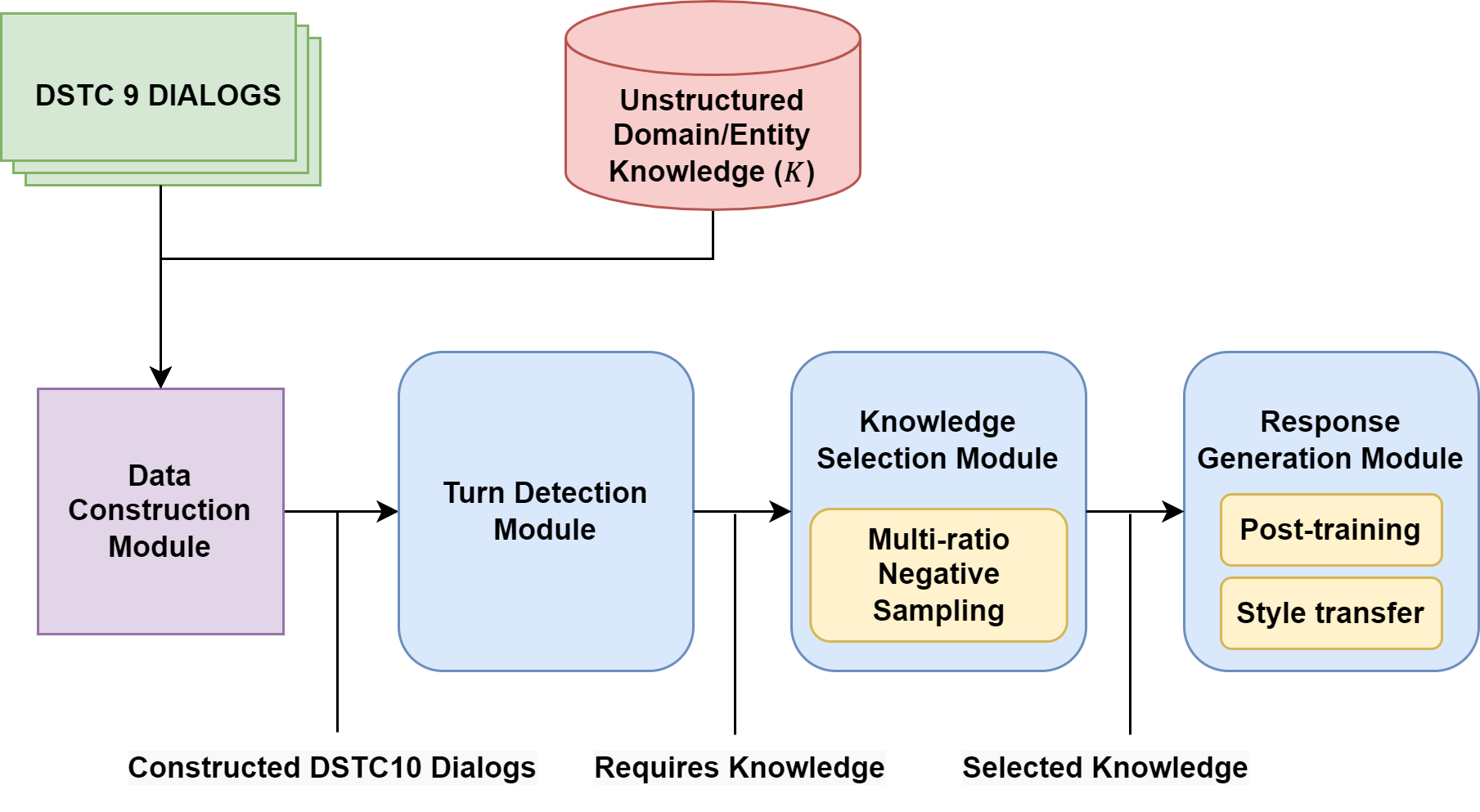}
\caption{Overall architecture of our knowledge-grounded task-oriented dialogue system.}
\label{overview}
\end{figure}

Recently, \texttt{DSTC9-Track1} \citep{dstc9} tackled this issue by proposing a new task that integrates external knowledge sources into TOD systems. By making the systems utilize frequently-asked questions (FAQs), which is a typical in-domain unstructured knowledge, we anticipate that TOD systems are able to respond to requests beyond the DB or API with no friction. Specifically, this track aims at developing a pipeline of three successive sub-tasks: 1) detecting whether external knowledge is required for a given dialog, 2) selecting a proper knowledge snippet from the entire knowledge, and 3) generating a response based on the retrieved knowledge. Various studies \citep{knover,mi2021towards,tang2021radge} have been conducted on unstructured knowledge-grounded TOD systems.

In reality, many TOD services include spoken conversation scenarios such as customer service and call centers. However, training such a system is much more difficult because spoken conversations contain speech recognition noise. In this respect, the Knowledge-grounded Task-oriented Dialogue Modeling on Spoken Conversations task or \texttt{DSTC10-Track2-Task2} has been proposed by \citet{DSTC10}. The task aims at the robustness of the systems against the gaps between written and spoken conversations. 

In this work, we build a knowledge-grounded TOD system that solves the \texttt{DSTC10} task. The outline of our system is shown in Figure \ref{overview}. First, we automatically construct new spoken conversation data using \texttt{DSTC9} conversation and \texttt{DSTC10} knowledge FAQ because the \texttt{DSTC10} task only provides new external knowledge and no spoken dialogues for training. And then, we use our synthetic dialogue data for training the detection, selection, and generation modules. Furthermore, we propose a new weighted negative sampling method in the selection module to improve. Finally, we apply the post-training method, which trains the pre-trained model on a large TOD corpus before fine-tuning, and the style transfer method, which learns style from responses similar to target system utterances in the generation step. This allows the system to generate a response appropriate for the dialogue contexts and similar to the target.

In summary, our contributions are as follows.
\begin{enumerate}
     \item \textbf{Automatic data construction:} We automatically construct new conversations required for model training using \texttt{DSTC9} data and new knowledge of \texttt{DSTC10}.
     \item \textbf{Weighted negative sampling:} We improve the model's performance by suggesting and applying a new weighted negative sampling in the selection module.
     \item \textbf{Post-training and style transfer:} We generate an appropriate response with a similar style to the target through the post-training and transfer process in the generation module.
\end{enumerate}

\section{Related Works}
The traditional pipeline approach \citep{WILLIAMS2007pipeline} in TOD systems consists of several subdivided components (NLU: natural Language understanding, DST: dialogue state tracking, DM: dialogue management, NLG: natural language generation). With the development of deep learning technology, TOD systems are gradually progressing from pipeline approach to end-to-end approach \citep{e2eDialogSystem,ham-etal-2020-end} that combines some or all components of an existing pipeline.

As the architecture of TOD systems has changed, a large body of TOD benchmarks has also been proposed. \citet{DSTC2-2014,WOZ} provide TOD datasets for restaurant domain. Recently, as large-scale multi-domain dataset such as MultiWOZ \citep{budzianowski-etal-2018-multiwoz2.0,eric2019multiwoz2.1} have emerged, research on multi-domain TOD are being actively conducted \citep{2019-TRADE,simpletod,UBAR}.

\texttt{DSTC9-Track1} \citep{dstc9} is a task that focuses on generating an appropriate response in a turn that requires external knowledge resources. Data is an augmented version of Multiwoz 2.1 \citep{eric2019multiwoz2.1} with out-of-API-coverage turns grounded on external knowledge sources beyond the original database entries. This task consists of three sub-tasks: turn detection, knowledge selection, and knowledge-grounded response generation. In this track, \citet{knover} applies the multi-scale (random, in-domain, in-entity, cross-entity) negative sampling and shows the best performance.
The second best system \citep{mi2021towards} improves the robustness of the system by introducing
data augmentation, joint task learning, and an error-fixing ensemble algorithm.

Style Transfer is a method that has been successfully employed in the vision field \citep{VISION}. It changes original data into data of the desired style. Since data for style transfer is expensive and difficult to obtain, augmentation methods are applied to construct data. In the field of natural language processing, a back-translation \citep{NMTBT, StyleTransferBT} method is widely used. This method converts the input to another language or another domain and then back to the original domain to secure the diversity of sentences. Specially back-translation is used not only in the natural language processing field but also in the automatic speech recognition (ASR) field as a method to acquire parallel text data for speech \citep{ASRBACK}.

\section{Problem Formalization}
\texttt{DSTC10-Track2-Task2} focuses on accessing external knowledge and generating appropriate answers, assuming that a conventional API-based TOD system already exists. Dialogue context is a sequence of $m$ utterances which is $C_i = \{u_1,s_2,...,u_m\}$, where $u_t$ and $s_t$ denote user utterance and system utterance respectively at the $t_{th}$ turn. A knowledge set $K \in \{k_1,...,k_n\}$ contains $n$ knowledge faq pairs $k_j$ composed of domain, entity, title and body. The final goal is to generate a knowledge reflected response $r_i$ by selecting the knowledge suitable for the conversation context when an utterance requires external knowledge. We define three sub-tasks as follows.

\begin{itemize}
\item\textbf{Detection:} Build a model to predict whether external knowledge access is needed to respond for a given dialogue context $C_i$. $y_i \in \{0,1\}$ denote truth label, $y_i=1$ indicate accessing external knowledge is required; otherwise, $y_i=0$. We define detection task formulation as follow:
\begin{equation}
    g_{detection}(C_i) \in \{0,1\}
\end{equation}

\item\textbf{Selection:} Select the most relevant knowledge for given the dialogue context $C_i$ and knowledge set $K$. The matching degree between $C_i$ and knowledge $k_j$ is denoted as $g_{selection}(C_i,k_j)$, and the knowledge with the highest matching degree becomes the related knowledge. we define selection task formulation as follow:
\begin{equation}
    g_{selection}(C_i,k_j) \in \{0,1\}
\end{equation}
where $y_i=1$ indicate $k_j$ is relevant with context $C_i$; otherwise, $y_i=0$.

\item\textbf{Generation:} Find a generative model $g_{generation}$ that generates a response $r_i$ suitable for given context $C_i$ and related knowledge $k_r$. We define generation task formulation as follow:
\begin{equation}
    g_{generation}(C_i,k_r)=r_i
\end{equation}
\end{itemize}

\section{Methodology}
\subsection{Automatic Data Construction}
In the \texttt{DSTC10} knowledge-grounded TOD task, no training conversation data are given. However, the existing \texttt{DSTC9} conversation data ($D_{9}$) is inappropriate for training because the knowledge of \texttt{DSTC9} ($K_9$) and \texttt{DSTC10} ($K_{10}$) is different and $D_{9}$ is not spoken conversation data. To address this issue, we developed a module that automatically constructs new \texttt{DSTC10} conversation data ($D_{10}$) based on $D_{9}$ and $K_{10}$.

To construct conversations about $K_{10}$, we replace the last user turn $u_m$ that requires knowledge from the $K_{9}$ to the new utterance that requires $K_{10}$. The steps are as follows: 1) create a conversation session template by removing the last user turn $u_m$, which requires knowledge, from the $D_{9}$. 2) select a knowledge snippet from $K_{10}$. 3) replace the dialogue template with the corresponding entity of the selected knowledge snippet, and substitute the last user-turn and target response with query and answer, which match the title and body of chosen knowledge snippet. Query and answer are the utterances with high similarity scores on knowledge title and body respectively from the candidate set; candidate set comprises $D_{9}$, $K_{9}$, the paraphrased \texttt{DSTC10} knowledge and $K_{10}$. We used sentence BERT \citep{sentencebert} to measure sentence similarity. We also train T5 \citep{T5} to paraphrase knowledge.
 
Finally, we intentionally add some ASR-like noises such as disfluencies and barge-ins into the generated conversations for imitating spoken conversations.
We employ a method similar to back translation among data augmentation methods to train the noise injector module. The learning process detail is as follows. First, we train a wav2vec2.0 \citep{wav2vec} based ASR model using the common voice dataset \citep{commonvoice} for data augmentation. Afterward, we train the BART \citep{bart} based noise injector module that transfers the written to spoken style, including ASR noise.

\subsection{Knowledge-seeking Turn Detection}
For a given dialogue context, we need to check whether it requires external knowledge or not. To address this problem, we approach this task to the binary classification task. We use GPT2 \citep{GPT2} base model, and the model input $x$ is as follows:
\begin{multline}
x = [BOS]\;[user]\;u_1\;[sys]\;s_2\;[user]\;u_3\;...\;u_m\;[EOS]
\end{multline}
where $[BOS], [EOS]$ are begin of the sentence token and the end of the sentence token, respectively. We also insert speaker tokens $[user], [sys]$ to distinguish user and system utterances. After that, the output state of the last token $T_{[LAST]}$ is used as the aggregate representation and passed through the single linear function as follow:
\begin{equation}
    g_{detection}(C_i)=\sigma(W_{detection}T_{[LAST]}+b)
\end{equation}
where $W_{detection}$ is a task-specific trainable parameter.
Eventually, the model weights are fine-tuned by using the cross-entropy loss function.
\begin{multline}
        Loss=-\sum\limits_{i}y_i \log(g_{detection}(C_i))\cr
                        +(1-y_i)\log(1-g_{detection}(C_i))
\end{multline}

\subsection{Knowledge Selection}
The system selects the appropriate knowledge from the entire knowledge set if the dialogue context requires external knowledge.  
We train the matching model $g_{selection}$ between the conversation history and each knowledge pair to select appropriate knowledge. 
We use a RoBERTa \citep{Roberta} base model for the knowledge selection. The input format $x$ of RoBERTa is as follows:
\begin{equation}
x = [CLS]\;u_1\;s_2\;u_3\;...\;u_m\;[SEP]\;k_j\;[EOS]
\end{equation}
where $[CLS], [SEP]$ are class token and separator token, respectively.
The final hidden vector of CLS token $T_{[CLS]}$ is used as the aggregate representation of the matching model and passed through the single linear function. If the given knowledge is related, the value is 1; if not related, the value is 0. 
Finally, the model is trained through cross-entropy loss for multi-class between related knowledge and negative samples as follow:
\begin{equation}
        Loss=-\sum\limits_{i} \sum\limits_{j}\;y_j \log(g_{selection}(C_i,k_j))
\end{equation}

\begin{table*}[t]
\resizebox{\textwidth}{!}{%
\begin{tabular}{c|ccc|ccc|cccccccc}
\toprule
\multicolumn{1}{c|}{Subtask} & \multicolumn{3}{c|}{Task1: Turn Detection} & \multicolumn{3}{c|}{Task2: Knowledge Selection} & \multicolumn{8}{c}{Task3: Response Generation}                           \\
\multicolumn{1}{c|}{Model}   & Precison       & Recall       & F1         & MRR@5        & Recall@1        & Recall@5       & BLEU-1 & BLEU-2 & BLEU-3 & BLEU-4 & METEOR & ROUGE-1 & ROUGE-2 & ROUGE-L \\ \hline
Baseline                     & 0.985          & 0.634        & 0.771      & 0.540        & 0.444           & 0.690          & 0.133  & 0.053  & 0.023  & 0.012  & 0.145  & 0.165   & 0.044   & 0.118   \\
Knover                       & 0.970          & 0.625        & 0.760      & 0.578        & 0.526           & 0.666          & 0.130  & 0.063  & 0.032  & 0.011  & 0.159  & 0.157   & 0.053   & 0.124   \\ \hline
\textbf{Ours}                         & \textbf{0.990}          & \textbf{0.961}        & \textbf{0.975}      & \textbf{0.831}        & \textbf{0.780}           & \textbf{0.887}          & \textbf{0.414}  & \textbf{0.327}  & \textbf{0.259}  & \textbf{0.194}  & \textbf{0.491}  & \textbf{0.492}   & \textbf{0.312}   & \textbf{0.469}  \\
\bottomrule
\end{tabular}
}
\caption{\label{table1} Experimental results on the validation set}
\end{table*}
\begin{table*}[t]
\resizebox{\textwidth}{!}{%
\begin{tabular}{c|ccc|ccc|cccccccc}
\toprule
\multicolumn{1}{c|}{Subtask} & \multicolumn{3}{c|}{Task1: Turn Detection} & \multicolumn{3}{c|}{Task2: Knowledge Selection} & \multicolumn{8}{c}{Task3: Response Generation}                           \\
\multicolumn{1}{c|}{Model}   & Precison       & Recall       & F1         & MRR@5        & Recall@1        & Recall@5       & BLEU-1 & BLEU-2 & BLEU-3 & BLEU-4 & METEOR & ROUGE-1 & ROUGE-2 & ROUGE-L \\ \hline
Baseline                     & 0.901          & 0.711        & 0.795      & 0.523        & 0.458           & 0.625          & 0.115  & 0.051  & 0.018  & 0.007  & 0.121  & 0.144   & 0.041   & 0.114   \\
Knover                       & 0.896          & 0.673        & 0.769      & 0.557        & 0.495           & 0.647          & 0.124  & 0.061  & 0.029  & 0.015  & 0.136  & 0.151   & 0.051   & 0.122   \\ \hline
\textbf{Ours (B12)}                         & \textbf{0.869}          & \textbf{0.938}        & \textbf{0.902}      & \textbf{0.732}        & \textbf{0.691}           & \textbf{0.79}           & \textbf{0.332}  & \textbf{0.235}  & \textbf{0.156}  & \textbf{0.098}  & \textbf{0.403}  & \textbf{0.384}   & \textbf{0.2}     & \textbf{0.374}  \\
\bottomrule
\end{tabular}
}
\caption{\label{table2} Experimental results on the test set}
\end{table*}
\begin{table}[t]
\resizebox{\linewidth}{!}{%
\begin{tabular}{c|ccc}
\toprule
Model            & Accuracy & Appropriateness & Average \\ \hline
Baseline: DSTC9  & 2.7425   & 2.7894          & 2.7659  \\
Baseline: Knover & 2.7793   & 2.7435          & 2.7614  \\ \hline
\textbf{Ours (B12)}        & \textbf{3.2546}   & \textbf{3.2336}          & \textbf{3.2441}  \\ \hline
Ground-truth     & 3.5769   & 3.4814          & 3.5292  \\
\bottomrule
\end{tabular}
}
\caption{\label{table3} Final human evaluation on the test set}
\end{table}

\subsubsection{Weighted Negative Sampling}
In general, random knowledge other than target knowledge is used as a negative sample when training a selection model. However, since most random negative samples are easily distinguished from target knowledge, the model has a problem learning only on easily distinguishable samples. In this respect, multi-scale negative sampling \citep{knover} train the model by classifying negative samples into multiple categories is proposed. However, we argue that the multi-scale negative sampling method overlooks that the difficulty of negative samples is different per category. 
To this end, we propose a weighted negative sampling method that trains the model more fine-grained manner than previous methods.
Our method gives different weight probabilities to each negative sample category to make the model focus on learning the negative samples that are difficult to distinguish. 

Negative sample categories in our method are as follows :\\
\textbf{Random:} knowledge randomly selected from the entire knowledge set. \\
\textbf{In-entity:} knowledge randomly chosen from among the knowledge in the same entity.\\
\textbf{In-domain:} knowledge chosen from among the knowledge in the same domain.\\
\textbf{Semantically-similar:} knowledge arbitrarily selected from among similar knowledge. The similarity between knowledge is obtained through BERT similarity \citep{sentencebert}.\\
Furthermore, the model may not train enough if the number of negative samples is inadequate for the weighted negative sampling method. Conversely, if there are many negative samplings, the false prediction of the model might increase. We set the appropriate number of negative samples to four through the experiments.

\subsection{Knowledge-grounded Generation}
Given the knowledge related to the conversation history, the system needs to generate an appropriate response. Specifically, the response should maintain coherency and context flow for conversation context and contain the user's information from the selected external knowledge. For this sub-task, we use GPT2 base as the model and input the conversation history and one related knowledge to the model as follows:
\begin{equation}
x = [BOS][know]\;k_{r}\;[user]\;u_1\;...\;u_m\;[EOS]
\end{equation}
where [know] is the knowledge tag to mark the start of knowledge and $k_r$ is the most related knowledge from the selection module.
The training objective for generation is the same with language modeling objective \citep{autoregressive}, which maximize next word prediction probability as follow:
\begin{equation}
        Loss=-\sum\limits_{i} \log(g_{generation}(r\vert C_i,k_r))
\end{equation}

\subsubsection{Post-training}
Since pre-trained models (PLMs) are trained with general corpus, the context representation for the specific task may be insufficient.
To address this issue, post-training methods \citep{dontstop,bertfp} that train PLMs once more with in-domain data before fine-tuning have been proposed. 
From this point of view, we post-train the model through the large TOD data, which contains generated \texttt{DSTC10} conversation, \texttt{DSTC9} conversation, \texttt{DSTC9} knowledge, \texttt{DSTC10} knowledge, and MultiWOZ. 
For post-training, we use the same training objective as pre-training, next-word prediction.

\subsubsection{Style Transfer}
When we fine-tune the generation task only with constructed \texttt{DSTC10} training data, generated responses have a different style with a response from validation data. Therefore the performance of the automatic evaluation metric decreases even if the accuracy or consistency of the response with dialogue context is well enough. To make the model generate a response that has a similar style to the validation set, we additionally train the model using an extra dataset for style transfer.

\section{Experiments}
\subsection{Settings}
We evaluated the model with the validation set and test set provided by \texttt{DSTC10-Track2-Task2}. As mentioned in the Automatic Data Construction Section, we use our automated constructed \texttt{DSTC10} dataset for training.

In the detection task, \texttt{DSTC10} training set, which does not apply the noise injection process, was used. Precision, recall, and F1 were used as evaluation metrics.
For the selection task, we trained the model through \texttt{DSTC10} train set. We measured performance through evaluation metrics such as $MRR@5$ \citep{MRR}, $recall@1$, and $recall@5$.
For post-training in the generation task, \texttt{DSTC9} dialogs, generate \texttt{DSTC10} dialogs, \texttt{DSTC10} knowledge FAQ, and MultiWOZ are used.
For style transfer, we learned the validation step and the test step with different data. For the validation step, generated \texttt{DSTC10} data and \texttt{DSTC9} test set were used, For test step, the \texttt{DSTC10} validation set was additionally used for training. We used $BLEU-1/2/3/4$ \citep{BLEU}, $Meteor$ \citep{meteor}, and $ROUGE-1/2/L$ \citep{rouge} for evaluation metric.

The \texttt{DSTC9} baseline \citep{dstc9} and the \texttt{DSTC9} top ranked model, $Knover$ \citep{knover}, were used as baseline. These models are trained with \texttt{DSTC9} conversation data and knowledge.

\subsection{Experimental Result}
Table \ref{table1} and Table \ref{table2} are the evaluation set and test set performance for each subtask.
Except for the precision of the detection task in the test set, our model shows significant improvement in performance for all metrics compared to the baseline.
The main reason is baseline models are trained with previous knowledge (\texttt{DSTC9} knowledge), while our model is trained with conversation data to reflect new knowledge (\texttt{DSTC10} knowledge).
In addition, new methods such as weighted negative sampling, post-training, and style transfer, improved the performance in the selection task and the generation task.
Table \ref{table3} shows the results of human evaluation. Our model shows enhanced performance in both accuracy and appropriateness compared to baselines.

\section{Further Analysis}
\begin{table}[t]
\centering
\begin{tabular}{c|ccc}
\toprule
Method            & MRR@5 & Recall@1 & Recall@5 \\ \hline
Random            & 0.77  & 0.682    & 0.878    \\
Multi-scale  & 0.783 & 0.712    & 0.887    \\ \hline
\textbf{Weighted (ours)} & \textbf{0.831} & \textbf{0.78}     & \textbf{0.887}  \\
\bottomrule
\end{tabular}
\caption{\label{table4} Comparison of weighted negative sampling method on validation set.}
\end{table}

\begin{table*}[t!]
\resizebox{\textwidth}{!}{%
\centering
\begin{tabular}{c|cccccccc}
\toprule
Method                        & BLEU-1 & BLEU-2 & BLEU-3 & BLEU-4 & METEOR & ROUGE-1 & ROUGE-2 & ROUGE-L \\ \hline
Baseline                      & 0.132  & 0.071  & 0.035  & 0.017  & 0.198  & 0.237   & 0.096   & 0.207   \\
+Style\_transfer              & 0.360  & 0.275  & 0.201  & 0.140  & 0.429  & 0.445   & 0.260   & 0.418   \\
\textbf{+Post-training+style transfer} & \textbf{0.414}  & \textbf{0.327}  & \textbf{0.259}  & \textbf{0.194}  & \textbf{0.491}  & \textbf{0.492}   & \textbf{0.312}   & \textbf{0.469}  \\
\bottomrule
\end{tabular}
}
\caption{\label{table5} Ablation study for generation task on the validation set.}
\end{table*}

\subsection{Effectiveness of Weighted Negative Sampling}
We compared multiple negative sampling techniques to investigate the effect of our weighted negative sampling method in Table \ref{table4}.
Random denotes the random selection of negative sampling from the whole knowledge. 
Multi-scale retrieves negative samples by dividing them into several categories (random, in-entity, in-domain, semantically-similar).
Weighted indicate a method of varying the probability of negative samples for each category. Weight for each category is manually set through the experiments. The ratio of the number of weighted negative samples in categories is 2:1:2:2 for random, in-entity, in-domain, and semantically similar.
As shown in Table \ref{table4}, a multi-scale negative sampling method is more effective than selecting random knowledge FAQ pairs as negative samples.
Moreover, our weighted negative sampling method, which learns with different ratios for each category, shows significant performance improvement.
This is because our method strengthens the ability to select relevant knowledge by seeing more difficult categories to distinguish.

\subsection{Effectiveness of Post-training and Style Transfer}
We have experimented to identify the effects of post-training and style transfer for the generation task in Table \ref{table5}.
Baseline is trained for the generation task without post-training and style transfer. It shows the lowest performance across automatic evaluation metrics because the style of system response in the \texttt{DSTC10} training data is different from the validation set.
On the other hand, the model with style transfer has significantly enhanced performance compared to baseline. It is because the model learns the responses style by training additional \texttt{DSTC9} test data, which has a similar response style.
The post-trained and style transferred model shows improved performance compared to the model that performed style transfer alone. This is because the model learns in-domain representation through post-training and is optimized for the TOD data distribution in advance.

\section{Conclusion}
In this work, we address \texttt{DSTC10-Track2-Task2}. To enhance the knowledge-grounded TOD system, we employ an automatic data construction method containing a noise injector module to generate spoken conversation data about \texttt{DSTC10} knowledge. We also propose weighted negative sampling to improve the knowledge selection model and apply post-training and style transfer for the generation task. The validity of our methods has been checked through experiments. As a final result, our approach ranked 6 on objective human evaluation.

We find the selection module has a significant effect on overall system performance. Therefore we plan to research the advanced knowledge selection model as future work.
\bibliography{aaai}
\bibliographystyle{aaai}
\end{document}